\documentclass[conference]{IEEEtran}
\usepackage[utf8]{inputenc}
\usepackage{graphicx}
\usepackage{rotating}
\usepackage{hyperref}
\ifCLASSINFOpdf
\else
\fi
\begin{document}
%
\title{Toxicity Detection for Indic Multilingual Social Media Content
}

\author{\IEEEauthorblockN{Manan Jhaveri \\ NMIMS University}

\and
\IEEEauthorblockN{Devanshu Ramaiya \\ NMIMS University}

\and
\IEEEauthorblockN{Harveen Singh Chadha \\ Thoughtworks}}


%


\maketitle

\begin{abstract}
Toxic content is one of the most critical issues for social media platforms today. India alone had 518 million social media users in 2020. In order to provide a good experience to content creators and their audience, it is crucial to flag toxic comments and the users who post that. But the big challenge is identifying toxicity in low resource Indic languages because of the  presence of multiple representations of the same text. Moreover, the posts/comments on social media do not adhere to a particular format, grammar or sentence structure; this makes the task of abuse detection even more challenging for multilingual social media platforms. This paper describes the system proposed by team 'Moj Masti' using the data provided by ShareChat/Moj in \emph{IIIT-D Multilingual Abusive Comment Identification} challenge. We focus on how we can leverage multilingual transformer based pre-trained and fine-tuned models to approach code-mixed/code-switched classification tasks. Our best performing system was an ensemble of XLM-RoBERTa and MuRIL which achieved a Mean F-1 score of 0.9 on the test data/leaderboard. We also observed an increase in the performance by adding transliterated data. Furthermore, using weak metadata, ensembling and some post-processing techniques boosted the performance of our system, thereby placing us 1st on the leaderboard.
\end{abstract}


%
\IEEEpeerreviewmaketitle

\section{Introduction}
Toxic comments refer to rude, insulting and offensive comments that can severely affect a person’s mental health and severe cases can also be classified as cyber-bullying. It often instills insecurities in young people, leading them to develop low self-esteem and suicidal thoughts. This abusive environment also dissuades people from expressing their opinions in the comment section which is supposed to be a safe and supportive space for productive discussions. Young children learn the wrong idea that adapting profane language will help them in seeking attention and becoming more socially acceptable. Hence, the task of flagging inappropriate content on social media is extremely important to create a healthy social space.

In this paper, we discuss how we leveraged AI to solve this task for us. This task is unprecedented because of the lack of useful data to train powerful models. We have used state-of-the-art transformer models pretrained on MLM task, fine-tuned with different architectures. Furthermore, we discuss some creative post-processing techniques that help to enhance the scores.

 

\section{Task Overview}
\subsection{Problem Formulation}
\emph{IIIT-D Multilingual Abusive Comment Identification} is an innovative challenge towards combating abusive comments on Moj, one of India's largest social media platform, in multiple regional Indian languages.
This paper is about a novel research problem focused on improving the social space for the members of the social media community. The focus area is improving abusive comment identification on social media in low-resource Indian languages.

There are multiple challenges that we need to combat while dealing with multilingual text data. Some of the main challenges in this task include:

\begin{itemize}
  \item There is a lack of resources about literature and grammar despite millions of native speakers using these languages. Building NLP algorithms with limited basic lexical resource is highly challenging. 
  \item Not all Indic languages fall into the same Linguistic Families. There's (1) Indo-Aryan which includes Hindi, Marathi, Gujarati, Bengali, etc (2) The Dravidian family consists of Tamil, Telugu, Kannada and Malayalam (3) Tribes of Central India speak Austric languages (4) Sino-Tibetan languages are spoken by tribes of the North Eastern India. 
  \item The posts/comments on social media do not adhere to a particular format, grammar or sentence structure. They are short, incomplete, filled with slangs, emoticons and abbreviations.
\end{itemize}

\subsection{Data Description}
In this challenge, the data provided was split into 2 parts: training data with 665k+ samples and test data with 74.3k+ samples. Key novelties around this dataset include:
\begin{itemize}
  \item Natural language comment text data in 13 Indic Languages labeled as Abusive(312k samples) or Not Abusive(352k samples). Fig. 1. tells us that Hindi is the most common language.
  \item The data is human annotated.
  \item Metadata based explicit feedback - post identification number, report count of the comment, report count of the post, count of likes on the comment and count of likes on the post.
\end{itemize}

\begin{figure}[htp]
    \centering
    \includegraphics[width=7.8cm]{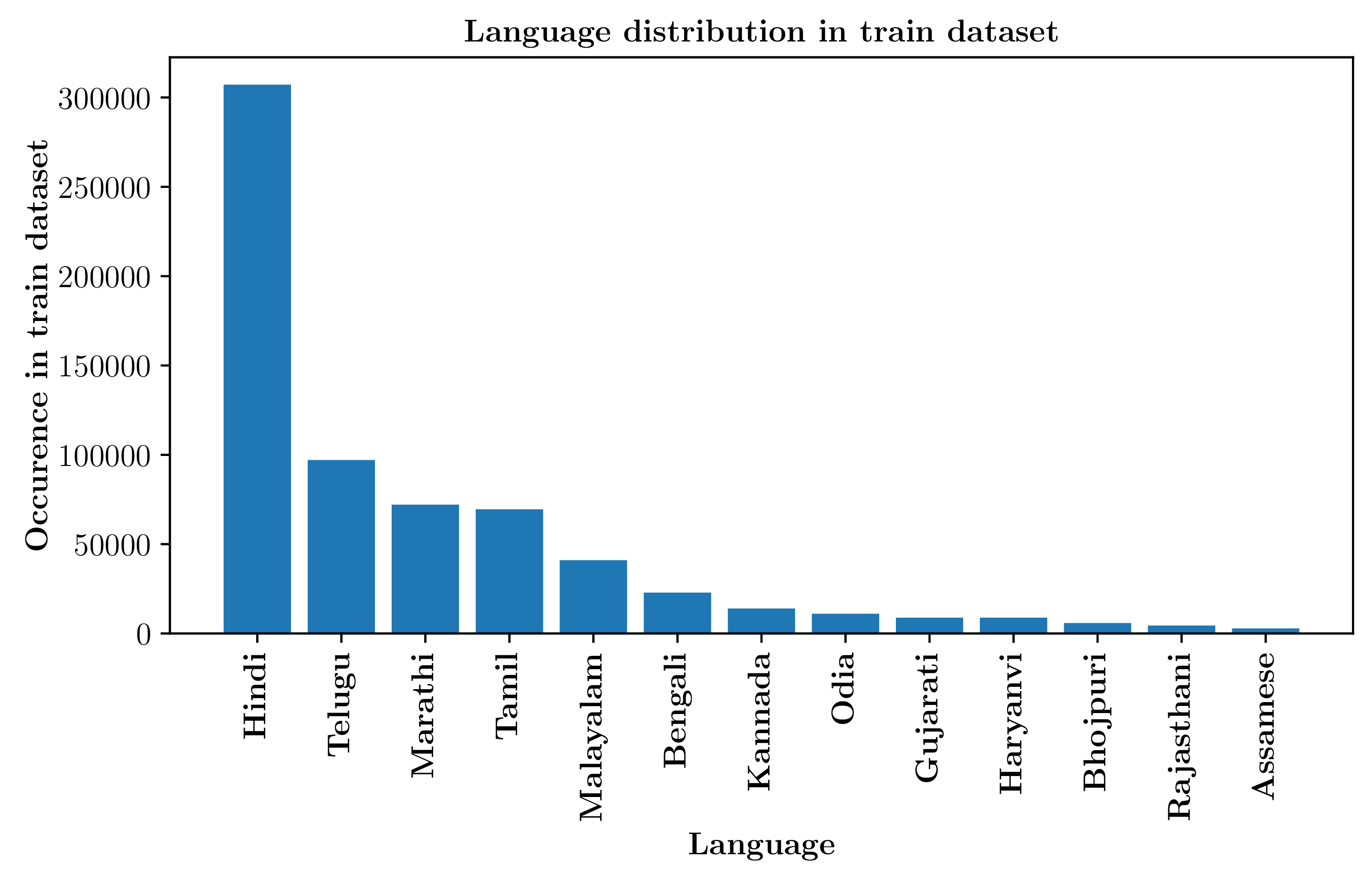}
    \caption{Language distribution in train dataset}
    \label{fig:lang_dist}
\end{figure}

\section{Model Building Approach and Evaluation}

After the research in Attention Mechanism \cite{vaswani2017attention} and the groundbreaking model - BERT \cite{devlin2019bert}, NLP space has been revolutionized and state-of-the-art Transformers have become the go-to option for almost all NLP tasks. For tackling this multilingual task, we chose the following models - 
\begin{itemize}
    \item \emph{XLM-RoBERTa} - It is a transformer-based masked language model trained on 100 languages, using more than two terabytes of filtered CommonCrawl data. \cite{conneau2020unsupervised}
    \item \emph{MuRIL} - A multilingual language model specifically built for Indic languages trained on significantly large amounts of Indian text corpora with both transliterated and translated document pairs, that serve as supervised cross-lingual signals in training. \cite{khanuja2021muril}
    \item \emph{mBERT} - MultilingualBERT (mBERT) is a transformer based language model trained on raw Wikipedia text of 104 languages. This model is contextual and its training requires no supervision - no alignment between the languages is done. \cite{k2020crosslingual}
    \item \emph{IndicBERT} - This model is trained on IndicCorp and evaluated on IndicGLUE. Similar to mBERT, a single model is trained on all Indian languages with the hope of utilizing the relatedness amongst Indian languages. \cite{kakwani-etal-2020-indicnlpsuite}
    \item \emph{RemBERT} - This model is pretrained on 110 languages using a Masked Language Modeling (MLM) objective. Its main difference with mBERT is that the input and output embeddings are not tied. Instead, RemBERT uses small input embeddings and large output embeddings. \cite{DBLP:conf/iclr/ChungFTJR21}
\end{itemize}

The Evaluation Metric for this task is Mean F1-score.

\subsection{Pre-training}
Pre-training using Masked Language Model(MLM) \cite{song2019mass} has been one of the most popular and successful methods in downstream tasks mostly due to their ability to model higher-order word co-occurrence statistics. \cite{sinha2021masked}

Since MuRIL is already a BERT encoder model pre-trained on Indic languages with MLM objective, we performed pre-training only on XLM-RoBERTa model (3 epochs for Base variant and 2 epochs for Large variant). We separated 10\% of the test data for evaluating the pre-train step. Table I shows the summary of this task. 
On average, the test F-1 score on downstream task was approximately 0.87 without performing MLM and 0.88 with MLM when we use the same settings to train the finetuned model in both the cases. This helped us to conclude that MLM on the given data certainly helps in the downstream tasks. 

\begin{table}[!t]
\renewcommand{\arraystretch}{1.3}
\caption{Validation loss for MLM pre-training}
\label{mlm_pretraining}
\centering
\begin{tabular}{|c|c|c|}
\hline
Info & XLM-RoBERTa Base & XLM-RoBERTa Large\\
\hline
Accelerator & Tesla P100 16GB & Tesla T4 16GB\\
\hline
Time Taken (hr) & 11.26 & 35.3 \\
\hline
Epoch 1 & 3.3819 & 3.0549\\
\hline
Epoch 2 & 3.3222 &  2.8466\\
\hline
Epoch 3 & 3.2946 & -\\
\hline
\end{tabular}
\end{table}

\subsection{Data Augmentation}
We performed data augmentation by adding transliterated data. We removed emojis from text and then we used uroman\footnote{\url{https://github.com/isi-nlp/uroman}} to generate additional transliterated data of 219114 samples. We also transliterated the test dataset and made the original plus transliterated dataset available on Kaggle  \footnote{\url{https://www.kaggle.com/harveenchadha/iitdtransliterated}}. 

\subsection{Model training}
Mainly, we used 2 different Model Architectures; one was the original architecture where we took the transformer outputs and found the probabilities and in the second one, we added a custom attention head for the transformer output before calculating the probabilities. 

We also experimented with a number of different truncation sizes of the input text length. We tried different models with truncation length of 64, 96, 128 and 256 and finally decided to go ahead with 128.

From Table II, we found out that MLM and Transliterated data had a positive impact on the performance. Moreover, a custom attention head boosted the score too. XLM-RoBERTa was the best performer followed by MuRIL. We also observed a slight difference between GPU and TPU accelerators -  models trained on GPU tend to perform much better. But since experimentation time on TPU was lower, we went ahead with it for most experiments. We used wandb\cite{wandb} to track most of our experiments, we are also releasing our experimentation logs \footnote{\url{https://wandb.ai/harveenchadha/iitd-private}}  

\begin{table*}[!t]
\renewcommand{\arraystretch}{1.3}
\caption{Model Experimentation Summary}
\label{model_exp_summary}
\centering
\begin{tabular}{|c|c|c|c|c|c|c|c|c|c|}
\hline
Id & Model & Data & Accelerator & Validation Strategy & CV & Test Score \\
\hline

1 & XLM-RoBERTa Base & Original & TPU v3-8 & 10\% split & 0.854 & 0.87674\\
\hline

2 & XLM-RoBERTa Large & Original & TPU v3-8 & 10\% split & 0.8601 & 0.8819 \\
\hline

3 & XLM-RoBERTa Base & Original + Transliterated & TPU v3-8 & 10\% split & 0.86 & 0.87989\\
\hline

4 & XLM-RoBERTa Large & Original + Transliterated & TPU v3-8 & 10\% split & 0.8626 & 0.88238 \\
\hline

5 & XLM-RoBERTa Large + MLM & Original & TPU v3-8 & 10\% split & 0.8631 & 0.88291 \\
\hline

6 & XLM-RoBERTa Large + MLM & Original + transliterated & TPU v3-8  & 10\% split & 0.863 & 0.88316 \\
\hline

7 & XLM-RoBERTa Base + MLM + Attention head & Original & Tesla P100 - 16GB & 4-Fold & 0.866075 & 0.8814\\
\hline

8 & XLM-RoBERTa Large + MLM + Attention head & Original + transliterated & RTX5000 24GB & 10\% split & 0.8669 & 0.88378 \\
\hline

9 & MuRIL Base & Original  & TPU v3-8 & 10\% split & 0.8520 & 0.87539\\
\hline

10 & MuRIL Large & Original & TPU v3-8 & 10\% split & 0.8546 & 0.87821 \\
\hline

11 & XLM-R Base + MLM + last 5 hidden states average & Original & Tesla P100 - 16GB & 4-Fold & 0.8662 & 0.88149 \\
\hline
12 & XLM-RoBERTa + MLM & Original + transliterated & TPU v3-8 & 4-Fold & 0.86825 & 0.8872\\
\hline
13 & XLM-RoBERTa Large & Original & TPU v3-8 & 10-Fold & 0.85759 & 0.8829 \\
\hline
14 &  RemBERT (Truncation Length = 64) & Original & Tesla P100 & 10\% split & 0.8529 & 0.877\\
\hline
15 & RemBERT & Original & TPU v3-8 & 10\% split & 0.8397 & 0.8678\\
\hline
16 &  mBERT & Original & TPU v3-8 & 10\% split & 0.8474 & 0.8724\\
\hline
17 & mBERT (Truncation Length = 64) & Original & Tesla P100 & 4-Fold & 0.8435 & 0.86327\\
\hline
18 & IndicBERT & Original & TPU v3-8 & 10\% split & 0.8306 & 0.8456\\
\hline
\end{tabular}
\end{table*}

As mentioned above, we also experimented with other transformer models but didn't achieve satisfactory results. RemBERT has outperformed XLM-RoBERTa in various tasks but underperformed in this case. mBERT also gave meagre results. Surprisingly, IndicBERT is the worst performer for this task.

Note - We transliterated the test data as well. So, whenever we have trained a model with transliterated data, the inference was done on the original text and transliterated text and the probabilities were combined with 7:3 ratio to form the final prediction.

\subsection{Ensembling}
We wanted to select diverse models so that each model makes different mistakes and by combining the learning of diverse models, we can get a better result. Hence, after a few experiments, we went forward with the following Ids from Table II -  2, 6, 8, 9, 12, 13.
We gave equal weights to all the models and our test  score was 0.88756. After changing the probability inference threshold to 0.55, the score increased to 0.88827. 

\subsection{Using Metadata}

As mentioned previously, we were also provided with weak metadata which was utilized in the following way - 

\subsubsection{Feature Engineering}
Due to the variation in timestamps, there were different values of reports and likes for the same post in different samples. 

For example, there is a post with \emph{post\_index} = 1, and it has 1 comment. When this comment was captured for the dataset, the like and report on the post were 5 and 10 respectively. After a few minutes, when a new comment was uploaded and captured for the dataset, the number of likes and reports had been changed to 10 and 20 respectively. Which means, now there are 2 samples with \emph{post\_index} = 1 in the dataset but the report count and like count recorded for them are different due to variation in the timestamp.

In order to deal with this, we created mean-value aggregated features and maximum-value aggregated features for report count and like count for posts as well as comments.

We also added length of characters in the comment, length of tokens in the comment and probabilities from the ensemble as a feature. We used this data to train boosting classifiers.

\subsubsection{Post-processing}
As per our findings, increasing the threshold gave us a boost and hence we decided to tweak the thresholds for each language. After experimenting with several thresholds, we found that the values in Table III gave the best score.

All the threshold values that we tested were multiples of 5. So, there must have been some edge cases that might be getting misclassified and in order to account for that, we increased the predicted probabilities by 0.01.

Table IV shows the performance of different boosting models. XGBoost and CatBoost were giving the highest score. We combined the probabilities from these models with XGBoost to CatBoost ratio of 6:4 and achieved the final best score of 0.90005.

\begin{table}[!t]
\renewcommand{\arraystretch}{1.3}
\caption{Language-wise Inference Thresholds}
\label{lang_thresh}
\centering
\begin{tabular}{|c|c|}
\hline
Language & Threshold\\
\hline
Marathi &  0.6\\
\hline
Malayalam &  0.5\\
\hline
Hindi &  0.54\\
\hline
Telugu &  0.6\\
\hline
Tamil &  0.5\\
\hline
Odia &  0.5\\
\hline
Gujarati &  0.4\\
\hline
Bhojpuri &  0.6\\
\hline
Haryanvi &  0.5\\
\hline
Assamese &  0.45\\
\hline
Kannada &  0.6\\
\hline
Rajasthani &  0.5\\
\hline
Bengali &  0.6\\
\hline
\end{tabular}
\end{table}

\begin{table}[!t]
\renewcommand{\arraystretch}{1.3}
\caption{Post-processing Models Summary}
\label{post_process_models}
\centering
\begin{tabular}{|c|c|c|}
\hline
Model & CV Average & Test Score\\
\hline
LightGBM &  0.9051 & 0.89933\\
\hline
XGBoost &  0.9051 & 0.89988\\
\hline
CatBoost &  0.9052 & 0.89980\\
\hline
\end{tabular}
\end{table}

\section{Conclusion}
In recent times, social media has become a hub for information exchange and entertainment. Results from this paper can be used to create systems that can flag toxicity and provide users with a healthier experience. Surprisingly, MuRIL which is trained specifically on Indian text did not perform well individually (both large and base variants) but it gave us an edge in the ensemble because it added to model diversity. For the same reason, we added non-pretrained models to the ensemble. The metadata in itself was very weak but combining it with the transformer output probabilities improved our predictions substantially. Even small tweaks like increasing the probability by 0.01 and language-wise inference thresholds had an astonishing impact in making a better system.


\section*{Acknowledgment}
We would like to thank Moj and ShareChat for providing the data and organizing the Multilingual Abusive Comment Identification Challenge along with IIIT-Delhi. We would also like to thank Kaggle for providing access to TPUs and jarvislabs.ai for providing credits to train the final model.



%

\bibliographystyle{plain}
\bibliography{mybib.bib}




\end{document}